\PassOptionsToPackage{unicode}{hyperref}
\PassOptionsToPackage{hyphens}{url}
\documentclass[
]{article}
\usepackage{amsmath,amssymb}
\usepackage{iftex}
\ifPDFTeX
  \usepackage[T1]{fontenc}
  \usepackage[utf8]{inputenc}
  \usepackage{textcomp} 
\else 
  \usepackage{unicode-math} 
  \defaultfontfeatures{Scale=MatchLowercase}
  \defaultfontfeatures[\rmfamily]{Ligatures=TeX,Scale=1}
\fi
\usepackage{lmodern}
\IfFileExists{kotex.sty}{\usepackage{kotex}}{}
\ifPDFTeX\else
\fi
\IfFileExists{upquote.sty}{\usepackage{upquote}}{}
\IfFileExists{microtype.sty}{
  \usepackage[]{microtype}
  \UseMicrotypeSet[protrusion]{basicmath} 
}{}
\makeatletter
\@ifundefined{KOMAClassName}{
  \IfFileExists{parskip.sty}{%
    \usepackage{parskip}
  }{
    \setlength{\parindent}{0pt}
    \setlength{\parskip}{6pt plus 2pt minus 1pt}}
}{
  \KOMAoptions{parskip=half}}
\makeatother
\usepackage{xcolor}
\usepackage{longtable,booktabs,array}
\usepackage{multirow}
\usepackage{cite}
\usepackage{float}
\usepackage{adjustbox}
\usepackage{calc} 
\usepackage{etoolbox}
\makeatletter
\patchcmd\longtable{\par}{\if@noskipsec\mbox{}\fi\par}{}{}
\makeatother
\IfFileExists{footnotehyper.sty}{\usepackage{footnotehyper}}{\usepackage{footnote}}
\makesavenoteenv{longtable}
\usepackage{graphicx}
\usepackage[margin=1in]{geometry}
\newcommand{\centerimage}[2][]{%
  \begin{center}
    \includegraphics[#1]{#2}
  \end{center}
}
\makeatletter
\def\maxwidth{\ifdim\Gin@nat@width>\linewidth\linewidth\else\Gin@nat@width\fi}
\def\maxheight{\ifdim\Gin@nat@height>\textheight\textheight\else\Gin@nat@height\fi}
\makeatother
\setkeys{Gin}{width=\maxwidth,height=\maxheight,keepaspectratio}
\makeatletter
\def\fps@figure{htbp}
\makeatother
\ifLuaTeX
  \usepackage{selnolig}  
\fi
\IfFileExists{bookmark.sty}{\usepackage{bookmark}}{\usepackage{hyperref}}
\IfFileExists{xurl.sty}{\usepackage{xurl}}{} 
\hypersetup{
  hidelinks,
  pdfcreator={LaTeX via pandoc}}

\title{\textbf{A Specialized Large Multimodal Model for \\ Interpreting PET/CT in Head and Neck Cancer}}
\author{%
\begin{minipage}{0.96\textwidth}
\centering
{\normalsize
\mbox{Haengbok Chung\textsuperscript{1,2}},
\mbox{SunGyu Kim\textsuperscript{2,3}},
\mbox{Joo hyun Lee\textsuperscript{2,4,5}}\\[0.25em]
\mbox{Sangjin Bae\textsuperscript{2,4,5}},
\mbox{Min Jeong Cho\textsuperscript{2,4,5}},
\mbox{Minseok Suh\textsuperscript{3,6}},
\mbox{Jae Sung Lee\textsuperscript{*,1,2,3,4,5,7,8}}
}\\[1.8em]
{\footnotesize
\begin{tabular}{@{}r@{\;}p{0.89\textwidth}@{}}
\textsuperscript{1} & Interdisciplinary Program in Artificial Intelligence, Seoul National University, Seoul, Republic of Korea.\\
\textsuperscript{2} & Department of Nuclear Medicine, Seoul National University College of Medicine, Seoul, Republic of Korea.\\
\textsuperscript{3} & Department of Nuclear Medicine, Seoul National University Hospital, Seoul, Republic of Korea.\\
\textsuperscript{4} & Interdisciplinary Program in Bioengineering, College of Engineering, Seoul National University Graduate School, Seoul, Republic of Korea.\\
\textsuperscript{5} & Integrated Major in Innovative Medical Science, Seoul National University Graduate School, Seoul, Republic of Korea.\\
\textsuperscript{6} & Department of Nuclear Medicine, Seoul National University Bundang Hospital, Seoul National University College of Medicine, Seongnam, Republic of Korea.\\
\textsuperscript{7} & Institute of Radiation Medicine, Medical Research Center, Seoul National University College of Medicine, Seoul, Republic of Korea.\\
\textsuperscript{8} & Brightonix Imaging Inc., Seoul, Republic of Korea.
\end{tabular}\\[0.55em]
\textsuperscript{*} For correspondence, contact Jae Sung Lee (jaes@snu.ac.kr).
}
\end{minipage}
}
\date{}

\begin{document}
\nocite{0,1,2,3,4,5,6,7,8,9,10,11,12,13,14,15,16,17,18,19,20,21,22,23,24,25,26,27,28,29,30,31,32,33}

\maketitle
\clearpage
\begin{center}
\LARGE\title{\textbf{A Specialized Large Multimodal Model for \\ Interpreting PET/CT in Head and Neck Cancer\\}}
\end{center}

\vspace{1cm}

\begin{abstract}

\noindent \textbf{Background:} Diagnosing head and neck cancer using PET/CT in nuclear medicine is clinically challenging and time-consuming due to the anatomical complexity of the region. This has led to a growing need for computer-aided diagnosis (CAD) systems that can facilitate faster and more accurate clinical decision-making. Recently, large multimodal models (LMMs) have emerged as a promising solution. However, state-of-the-art generalist models still face several limitations in medical contexts, such as insufficient domain-specific knowledge, privacy and security concerns, and verbosity. Consequently, standalone specialized LMMs are receiving increasing attention as a next-generation approach to CAD.

\noindent \textbf{Purpose:} This paper demonstrates the feasibility of using a specialized Large Multimodal Model (LMM) to support the diagnosis of head and neck cancer by automatically interpreting PET/CT images. This was achieved through the construction of a large-scale, multi-institutional dataset, the design of a tailored training curriculum, and the adoption of autoregressive training strategies. The model was evaluated across four external institutions with diverse imaging devices and consistently outperformed state-of-the-art generalist models such as ChatGPT.

\noindent \textbf{Methods:} A baseline generalist LMM (LLaVA-NeXT) was fine-tuned using a two-level curriculum. To train the model, we constructed a large-scale dataset of image--conversation pairs enriched with clinically important annotations---such as the presence of a primary tumor and the location of metastatic lymph nodes---curated by two radiologists using publicly available data sources. In Level 1, the model was trained on 28,000 image--conversation pairs to learn basic but essential information, including modality type and the presence of hypermetabolism. In Level 2, using 12,975 image--conversation pairs, the model was further trained to detect diagnostically critical features such as the presence of a primary tumor and the existence and anatomical location of lymph node metastases.

\noindent \textbf{Results:} The proposed LMM exhibited substantially superior performance compared to state-of-the-art generalist models such as ChatGPT and LLaVA-NeXT, based on ROUGE-L, ROUGE-S, Cosine Similarity, Precision, Recall, and F1 scores. In the Level-2 external validation, our model achieved scores of 0.8751, 0.8794, 0.8324, 0.8794, 0.8711, and 0.8751 across the respective metrics, whereas the generalist models consistently scored below 0.1. To assess the model's ability to capture clinically meaningful details---specifically regarding primary tumor and cervical lymph node metastasis---we further isolated and evaluated only the primary tumor and lymph-node--related segments from the generated diagnostic sentences. In this focused analysis, the model achieved accuracies of 83.14 ± 1.15\% in internal validation and 69.03 ± 0.81\% in external validation for classifying the existence of primary tumor. For the localizing lymph node metastasis, our model scored 0.6389, 0.6257, 0.5287, 0.5782, 0.6371, and 0.6648 respectively, across the same evaluation metrics, underscoring its potential clinical utility in real-world diagnostic scenarios.

\noindent \textbf{Conclusion:} We assessed the feasibility of utilizing a specialized Large Multimodal Model (LMM) to facilitate rapid and accurate diagnosis, as well as to support medical education. Quantitative evaluations reveal that the specialized LMM significantly outperforms state-of-the-art generalist models such as ChatGPT. These findings highlight the potential of domain-specific LMMs as a promising direction for clinical translation.\\


\end{abstract}

\pagebreak

\section{Introduction}

Head and neck cancers \cite{0} arise usually from the squamous cells lining the mucosal surfaces of the oral cavity, pharynx, larynx, and adjacent structures. The timely and precise assessment of the primary tumor and regional lymph node involvement is critical for TNM staging, therapeutic decision-making and clinical outcome prediction. However, due to the anatomical complexity of the head and neck region, reliable evaluation of tumor extent and nodal involvement still remains challenging \cite{1}.

\centerimage[width=6.5in,height=1.06042in]{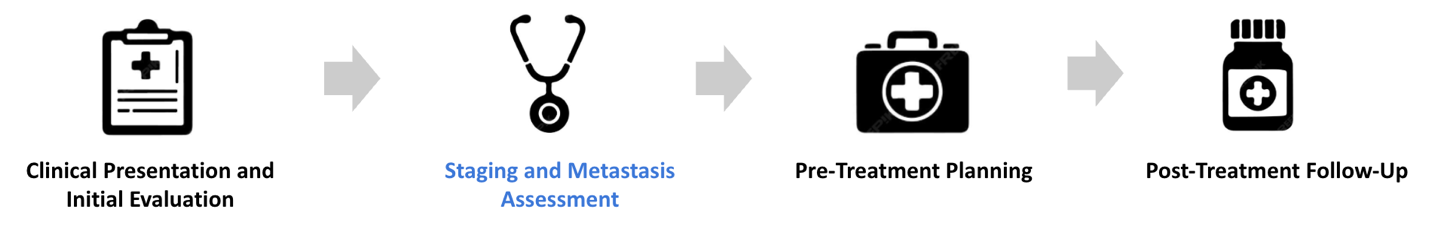}

Figure 1. The diagnostic workflow in nuclear medicine. The steps highlighted in blue indicate the key stages where our proposed LMM can be integrated to provide diagnostic assistance.

Positron Emission Tomography/Computed Tomography (PET/CT) has become an essential modality in head and neck cancer management by providing complementary metabolic and anatomical information. This dual-modality imaging plays a pivotal role in initial staging, metastasis assessment, pre-treatment planning, and post-treatment response evaluation. In particular, 2-deoxy-2-{[}\textsuperscript{18}F{]}fluoro-D-glucose ({[}\textsuperscript{18}F{]}FDG) PET/CT enables both anatomic localization and metabolic characterization of tumors and nodes, and is widely adopted for staging and treatment monitoring \cite{2,3,4}.

Despite its established clinical value, expert interpretation of PET/CT remains limited worldwide due to the shortage of trained nuclear medicine physicians \cite{5,6,7}. This scarcity of expertise may delay the diagnostic and therapeutic pathway and increase the risk of interpretive errors, ultimately impacting patient management. Therefore, to mitigate these limitations, it is necessary to develop assistive technologies that can augment clinical expertise and broaden access to high-quality diagnostic services.

Large multimodal models (LMMs) have recently emerged as promising approaches for supporting medical image interpretation and diagnostic decision-making, although their clinical integration remains at an early stage of development \cite{8,9,10}. General-purpose LMMs, such as ChatGPT \cite{11}, face substantial barriers to clinical adoption due to insufficient domain knowledge, privacy concerns, high computational demands, and variability of outputs \cite{12,13}. These limitations underscore the necessity for domain-specific models tailored to medical imaging. Therefore, specialized LMM development has been particularly active for medical images, such as chest X-rays \cite{14,15}. In contrast, LMMs for PET/CT remain limited, largely due to the scarcity of large-scale datasets and the unique multimodal characteristics of PET/CT. Nevertheless, advanced PET/CT LMMs hold substantial potential because of their strong ability to integrate multimodal information and interpret complex images.

Motivated by this potential, in this study, we developed and evaluated a specialized LMM for the interpretation of {[}\textsuperscript{18}F{]}FDG PET/CT images of head and neck cancer. The model was designed to support two tasks: (i) detection of primary tumors and metastatic lymph nodes, and (ii) interpretation of key PET/CT findings, such as hypermetabolic regions. To ensure robustness and generalizability, the model was trained on a large-scale, multi-center, multi-instrument dataset established through ongoing collaborative efforts in our society. Furthermore, we adopted a specialized two-level curriculum learning strategy \cite{24} to optimize the model to support head and neck cancer PET/CT interpretation. This progressive framework enables the model to acquire fundamental imaging concepts before advancing to clinically salient diagnostic inference steps, thereby enhancing its accuracy and reliability.

\section{Materials and Methods}

\centerimage[width=6.5in,height=2.29514in]{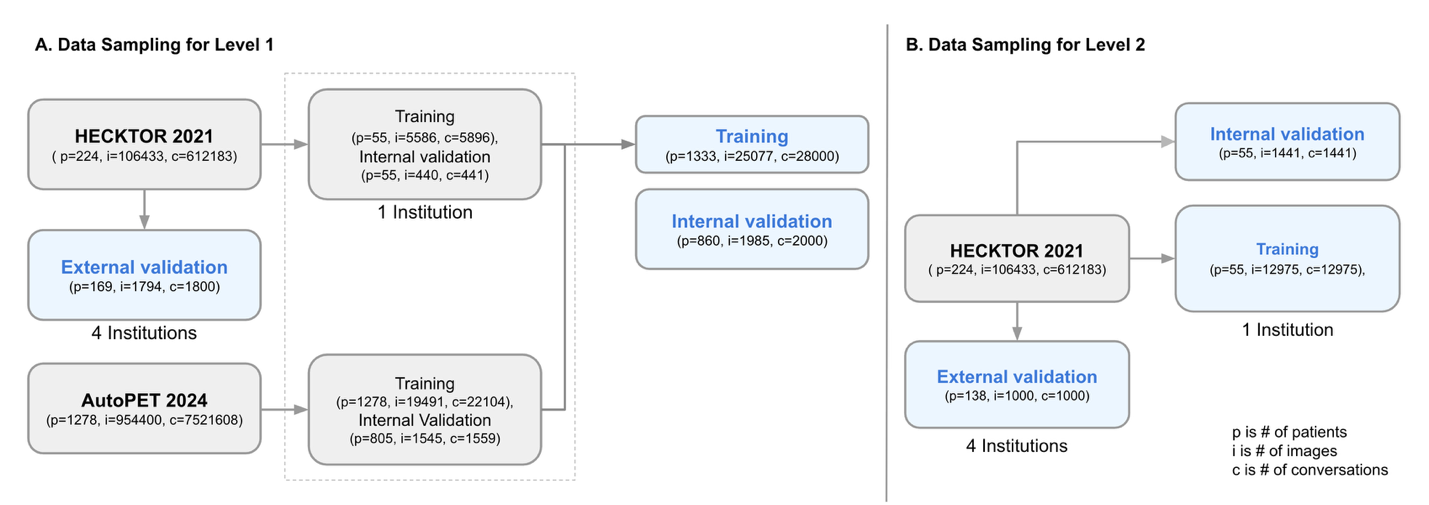}

Figure 2. \textbf{Flowchart of the data selection process for LMM training and evaluation at Level 1 and Level 2.} The curated dataset was randomly partitioned into training, internal validation, and external validation sets. This partitioning was stratified to ensure a balanced distribution of key data and patient characteristics (e.g., imaging modality, view, gender), as detailed in Table 1-4. A comprehensive description of the dataset construction is provided in the Supplementary Materials.

\begin{table}[H]
\centering
\small
\renewcommand{\arraystretch}{0.93}
\setlength{\tabcolsep}{4pt}
\begin{tabular}{@{}llrrrr@{}}
\toprule
\textbf{Characteristic} & \textbf{Type} & \shortstack{\textbf{Training}\\\textbf{(c = 28,000)}} & \shortstack{\textbf{Internal}\\\textbf{Validation}\\\textbf{(c = 2,000)}} & \shortstack{\textbf{External}\\\textbf{Validation}\\\textbf{(c = 1,800)}} & \emph{\textbf{P Value}} \\
\midrule
\multirow{3}{*}{Modality} & PET & 11,011 (39.3) & 799 (40.0) & 890 (49.4) & \multirow{3}{*}{\textless0.001} \\
& PET/CT & 9,125 (32.6) & 655 (32.8) & 540 (30.0) \\
& CT & 7,864 (28.1) & 546 (27.3) & 370 (20.5) \\
\midrule
\multirow{3}{*}{View} & Axial & 12,799 (45.7) & 936 (46.8) & 613 (34.0) & \multirow{3}{*}{\textless0.001} \\
& Sagittal & 7,513 (26.8) & 527 (26.4) & 497 (27.6) \\
& Coronal & 7,688 (27.5) & 537 (26.8) & 690 (38.3) \\
\midrule
\multirow{2}{*}{Hypermetabolism} & Exist & 15,954 (57.0) & 1,129 (56.4) & 739 (41.0) & \multirow{2}{*}{\textless0.001} \\
& Not Exist & 12,046 (43.0) & 871 (43.6) & 1,061 (58.9) \\
\bottomrule
\end{tabular}
\caption{\footnotesize Characteristics of the Datasets for Level 1 Training, Internal Validation, and External Validation.}
\label{tab:1}
\end{table}

\begin{table}[H]
\centering
\small
\renewcommand{\arraystretch}{0.93}
\setlength{\tabcolsep}{4pt}
\begin{tabular}{@{}llrrrr@{}}
\toprule
\textbf{Characteristic} & \textbf{Type} & \shortstack{\textbf{Training}\\\textbf{(p = 1,333)}} & \shortstack{\textbf{Internal}\\\textbf{Validation}\\\textbf{(p = 860)}} & \shortstack{\textbf{External}\\\textbf{Validation}\\\textbf{(p = 169)}} & \emph{\textbf{P Value}} \\
\midrule
Age & - & 65 (55-74) {[}11-95{]} & 65 (56-75) {[}17-95{]} & 63 (57-70) {[}34-90{]} & 0.241 \\
\midrule
\multirow{3}{*}{Gender} & Male & 543 (40.7) & 340 (39.5) & 124 (73.3) & \multirow{3}{*}{\textless0.001} \\
& Female & 408 (30.6) & 240 (27.9) & 45 (26.6) \\
& Unknown & 382 (28.7) & 280 (32.6) & 0 (0) \\
\midrule
\multirow{3}{*}{Anatomic Site} & Partial Body & 87 (6.5) & 76 (8.8) & 137 (81.0) & \multirow{3}{*}{\textless0.001} \\
& Whole Body & 1229 (92.2) & 774 (90.0) & 32 (18.9) \\
& Unprovided & 17 (1.3) & 10 (1.2) & 0 (0) \\
\midrule
\multirow{6}{*}{Disease} & Head and Neck Cancer & 55 (4.1) & 55 (6.4) & 169 (100) & \multirow{6}{*}{\textless0.001} \\
& Lung Cancer & 167 (12.5) & 100 (11.6) & 0 \\
& Prostate Cancer & 355 (26.6) & 261 (30.3) & 0 \\
& Melanoma & 169 (12.7) & 108 (12.6) & 0 \\
& Lymphoma & 135 (10.1) & 76 (8.8) & 0 \\
& Negative & 452 (33.9) & 260 (30.2) & 0 \\
\bottomrule
\end{tabular}
\caption{\footnotesize Patient Characteristics of the Level 1 Training, Internal Validation, and External Validation Cohorts.}
\label{tab:2}
\end{table}

\begin{table}[H]
\centering
\small
\renewcommand{\arraystretch}{0.93}
\setlength{\tabcolsep}{4pt}
\begin{tabular}{@{}llrrrr@{}}
\toprule
\textbf{Characteristic} & \textbf{Type} & \shortstack{\textbf{Training}\\\textbf{(c = 12,975)}} & \shortstack{\textbf{Internal}\\\textbf{Validation}\\\textbf{(c = 1,441)}} & \shortstack{\textbf{External}\\\textbf{Validation}\\\textbf{(c = 1,000)}} & \emph{\textbf{P Value}} \\
\midrule
\multirow{3}{*}{Modality} & PET & 6,515 (50.1) & 713 (49.4) & 495 (55.8) & \multirow{3}{*}{\textless0.001} \\
& PET/CT & 4,768 (36.7) & 538 (37.3) & 378 (42.6) \\
& CT & 1,692 (13.0) & 190 (13.1) & 13 (1.4) \\
\midrule
\multirow{3}{*}{View} & Axial & 2,556 (19.6) & 306 (21.2) & 257 (25.7) & \multirow{3}{*}{\textless0.001} \\
& Sagittal & 5,268 (40.5) & 572 (39.6) & 310 (31.0) \\
& Coronal & 5,151 (39.7) & 563 (39.0) & 433 (43.3) \\
\midrule
\multirow{2}{*}{Primary Tumor} & Exist & 2,089(16.1) & 233 (16.1) & 305 (30.5) & \multirow{2}{*}{\textless0.001} \\
& Not Exist & 10,886 (83.9) & 1208 (83.8) & 695 (69.5) \\
\midrule
\multirow{2}{*}{Lymph Node} & Exist & 9,375 (72.2) & 1041 (72.2) & 900 (90.0) & \multirow{2}{*}{\textless0.001} \\
& Not Exist & 3,600 (27.7) & 400 (27.7) & 100 (10.0) \\
\bottomrule
\end{tabular}%
\caption{\footnotesize Characteristics of the Datasets for Level 2 Training, Internal Validation, and External Validation.}
\label{tab:3}
\end{table}

\begin{table}[H]
\centering
\small
\renewcommand{\arraystretch}{0.93}
\setlength{\tabcolsep}{4pt}
\begin{tabular}{@{}llrrrr@{}}
\toprule
\textbf{Characteristic} & \textbf{Type} & \shortstack{\textbf{Training}\\\textbf{(p = 55)}} & \shortstack{\textbf{Internal}\\\textbf{Validation}\\\textbf{(p = 55)}} & \shortstack{\textbf{External}\\\textbf{Validation}\\\textbf{(p = 138)}} & \emph{\textbf{P Value}} \\
\midrule
Age & - & 62 (56-66) {[}45-81{]} & 62 (56-66) {[}45-81{]} & 63 (57-70) {[}34-90{]} & 0.399 \\
\midrule
\multirow{2}{*}{Gender} & Male & 43 (78.1) & 43 (78.1) & 101 (73.1) & \multirow{2}{*}{0.662} \\
& Female & 12 (21.8) & 12 (21.8) & 37 (26.8) \\
\midrule
\multirow{2}{*}{Anatomic Site} & Partial Body & 55 (100) & 55 (100) & 111 (80.4) & \multirow{2}{*}{\textless0.001} \\
& Whole Body & 0 (0) & 0 (0) & 27 (19.5) \\
\bottomrule
\end{tabular}
\caption{\footnotesize Patient Characteristics of the Level 2 Training, Internal Validation, and External Validation Cohorts.}
\label{tab:4}
\end{table}

\subsection{Dataset}

For model development, we constructed a large-scale dataset of image-conversation pairs, where each image was paired with a question and an expert-provided answer, curated from publicly available data sources. Two nuclear medicine experts performed detailed annotations of clinically relevant findings, including the presence and location of primary tumors and metastatic lesions. A full description of the dataset construction process is provided in the Supplementary Notes.

This retrospective study used multi-center {[}\textsuperscript{18}F{]}FDG PET/CT data obtained from two international deep learning challenges: HECKTOR2021 \cite{16} and AutoPET2024 \cite{17}. The training cohort included head and neck PET/CT images from an institution in HECKTOR2021 (Hôpital Général Juif, Montréal, CA) and whole-body PET/CT images from two institutions in AutoPET2024 (University Hospital Tübingen, Tübingen, Germany, and University Hospital of the LMU, Munich, Germany). To ensure scanner heterogeneity, the images were acquired using various PET/CT scanners, including the GE Discovery ST, Siemens Biograph 64-4R TruePoint, Siemens Biograph mCT Flow 20, and GE Discovery 690. From this aggregated dataset, we randomly sampled 28,000 image-conversation pairs from 1,333 patients for the Level 1 training set, and 12,975 pairs from 55 patients for the Level 2 training set (Figure 2).

For model evaluation, we composed validation cohorts using unseen data from four independent institutions in HECKTOR2021: Centre Hospitalier Universitaire de Sherbrooke (Sherbrooke, CA), Centre Hospitalier de l'Université de Montréal (Montréal, CA), Centre Hospitalier Universitaire de Poitiers (Poitiers, France), and Hôpital Maisonneuve-Rosemont (Montréal, CA). These hold-out datasets also featured scanner heterogeneity, with images acquired using the Philips Gemini GXL 16, GE Discovery STE, and Siemens Biograph mCT 40. From these data, we created distinct internal and external validation sets for both levels. For Level 1, we allocated 2,000 image-conversation pairs (from 860 patients) for internal validation and 1,800 pairs (from 169 patients) for external validation. For Level 2, the internal validation set comprised 1,441 pairs (from 55 patients), while the external validation set contained 1,000 pairs (from 138 patients).

All 3D images underwent a standardized preprocessing pipeline \cite{16} that included resampling, intensity clipping, registration, and normalization to a {[}0, 1{]} range. From these, 2D slices were extracted and uniformly resized to 256 × 256 × 3 for model input. The resulting dataset was diverse, incorporating multiple imaging modalities (PET, PET/CT, CT), three anatomical views (axial, sagittal, coronal), and varied scan coverage (partial-body and whole-body). The clinical composition also differed across cohorts: the Level 1 training and internal validation cohorts included patients with various malignancies (head and neck cancer, lung cancer, prostate cancer, melanoma, and lymphoma), while the Level 1 external validation and all Level 2 cohorts consisted exclusively of patients with head and neck cancer. A summary of data and patient characteristics is provided in Tables 1--4, with proportions for each characteristic shown in parentheses.

\centerimage[width=6.5in,height=2.78125in]{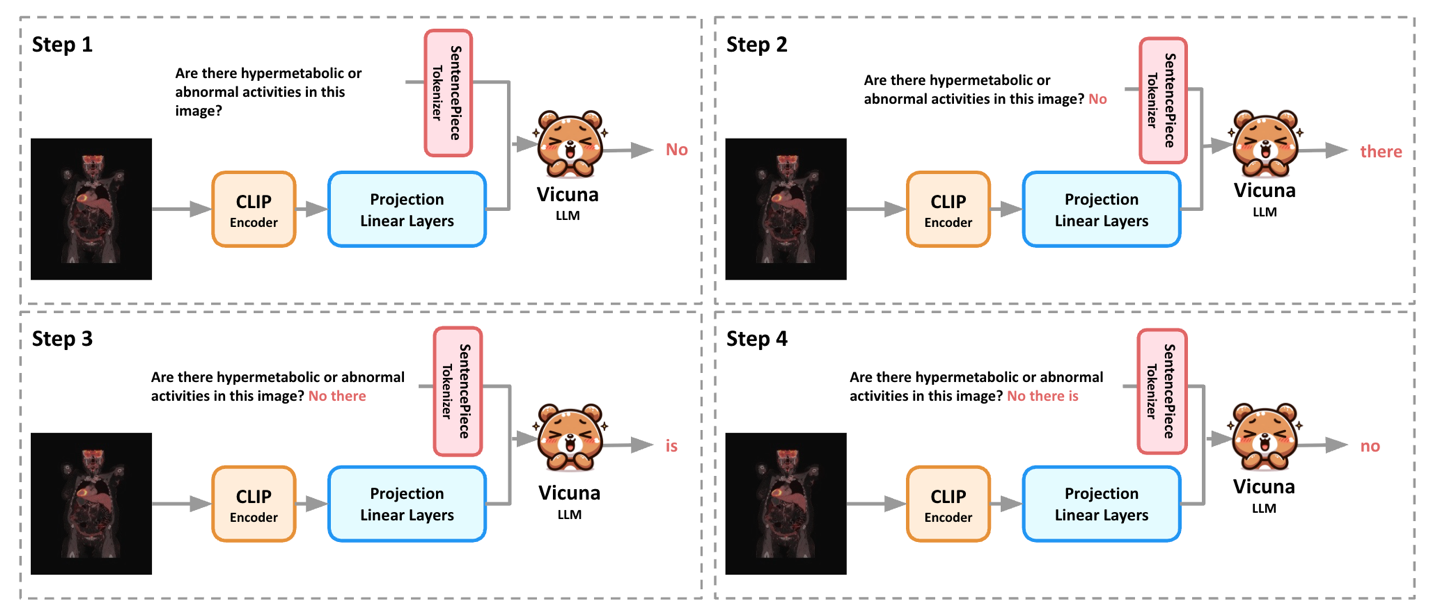}

Figure 3. The baseline model (LLaVA-NeXT) was trained and evaluated using an autoregressive generation method, as opposed to teacher forcing. In this procedure, the model sequentially predicts the next token in an answer by conditioning on the input question combined with all previously generated tokens.

\subsection{Network Model}

Our baseline LMM was LLaVA-NeXT \cite{18}, a 7-billion parameter model featuring a CLIP \cite{19} vision encoder and the Vicuna \cite{20,21} Large Language Model. As illustrated in Figure 3, the model processes an image and a question to generate a corresponding answer. We employed Low-Rank Adaptation (LoRA) \cite{22} for efficient fine-tuning, using the AdamW \cite{23} optimizer with a learning rate of 1e-5. The LoRA-specific hyperparameters were a rank of 128, an alpha of 128, and a dropout rate of 0.1, while text generation was controlled with a temperature of 2.0 and a top-p of 1.0.

\subsection{Network Training and Loss Function}

To optimize the model for diagnostic support in head and neck cancer, we adopted a specialized two-level curriculum learning strategy \cite{24}, as mentioned previously. Training commences at Level 1, where the model learns from foundational questions that are simpler but conceptually prerequisite to the final task. At Level 2, the curriculum advances to more complex and clinically salient diagnostic questions designed to foster advanced reasoning. This progressive approach enables the model to consolidate its understanding of basic concepts before addressing more challenging clinical scenarios.

More specifically, at Level 1, the model is asked four different questions: (1) modality classification (``Among PET, PET/CT, CT, and MRI, what type of scan is this?''), (2) lesion detection (``Are there any hypermetabolic lesions or abnormalities in this image?''), (3) anatomical view identification (``Among sagittal, coronal, and axial views, what is the view of the image?''), and (4) contrast agent usage (``Was a contrast agent used in this image?''). At Level 2, the task is escalated to clinically relevant diagnostic queries specific to head and neck cancers. Given that patients are suspected of having head and neck cancer, the model is asked a specific question: ``This image is a slice from an FDG PET/CT scan performed on a patient with oropharyngeal cancer. Please assess whether a primary tumor or any cervical lymph nodes suspicious for metastasis are visible in this slice. If any lymph nodes are visible, please specify the station (IL, IR, IIL, IIR, IIIL, IIIR, IVL, IVR, VL, VR, VIL, VIR, VIIL, VIIR, VIIIL, VIIIR, none).''

We diverged from the default teacher-forcing paradigm of LLaVA-NeXT by fine-tuning our model using a fully autoregressive approach \cite{25}. This strategy was deliberately chosen to mitigate the discrepancy between the training and inference environments. By conditioning the model on its own previously generated outputs during training, we simulated real-world inference conditions more faithfully, thereby improving both the stability and reliability of model performance. 

When \(L_{p}\) indicates the predicted answer logits and \(L_{l}\) the ground-truth logits, we add end-of-sequence (EOS) padding to the shorter side as shown in the following equations.

\begin{center}
\(L_{p}^{'}\lbrack i\rbrack = \left\{ \begin{array}{r}
L_{p}\lbrack i\rbrack\ (\text{if}\ i \leq n_{p}) \\
{EOS}_{c}\ ({\text{if}\ n}_{p} < i \leq n_{m})
\end{array} \right.\ \) (1)

\(L_{l}^{'}\lbrack i\rbrack = \left\{ \begin{array}{r}
L_{l}\lbrack i\rbrack\ (\text{if}\ i\  \leq n_{l}) \\
{EOS}_{c}\ ({\text{if}\ n}_{l} < i \leq n_{m})
\end{array} \right.\ \) (2)
\end{center}

where \(n_{p}\) and \(n_{l}\) are the sequence lengths of the predicted and ground-truth logits, respectively, and \(n_{m} = \max(n_{p}, n_{l})\). \({EOS}_{c}\) is a vector in which the index of the EOS token has a value of 0.999 and all remaining elements have small values. We apply cross-entropy loss between \(L_{p}^{'}\) and \(L_{l}^{'}\).

\subsection{Evaluations and Statistical Analyses}

We conducted a comprehensive quantitative evaluation of the model's performance on both Level 1 and Level 2 tasks using the internal and external validation datasets. Model outputs were assessed with multiple text generation metrics: ROUGE-L \cite{26}, ROUGE-2 \cite{26}, cosine similarity, precision, recall, and F1 score. For clinically focused evaluation, we specifically analyzed the model's predictions for primary tumor presence (derived from the first output token) and lymph node metastasis (derived from the final five tokens). The accuracy of lymph node localization was further assessed by calculating F1 scores for each individual nodal station.

All model performance metrics were calculated across three independent runs with different random seeds (42, 84, and 200) and are reported as the mean ± standard deviation. To compare the characteristics of the study cohorts, statistical tests were performed among the training, internal validation, and external validation sets. The chi-square ($\chi^2$) test \cite{27} was used for categorical variables, while the Kruskal--Wallis test \cite{28} was used to analyze differences in patient age due to its non-normal distribution. For all analyses, a p-value \textless{} .05 was considered statistically significant.

\section{Results}

\centerimage[width=5.14579in,height=2.36893in]{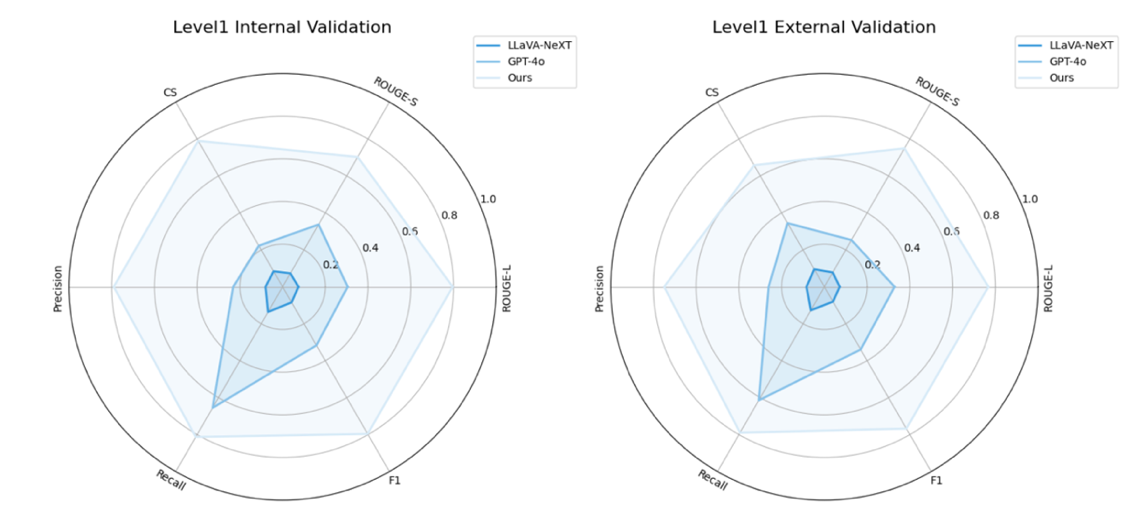}

\centerimage[width=5.15534in,height=2.35736in]{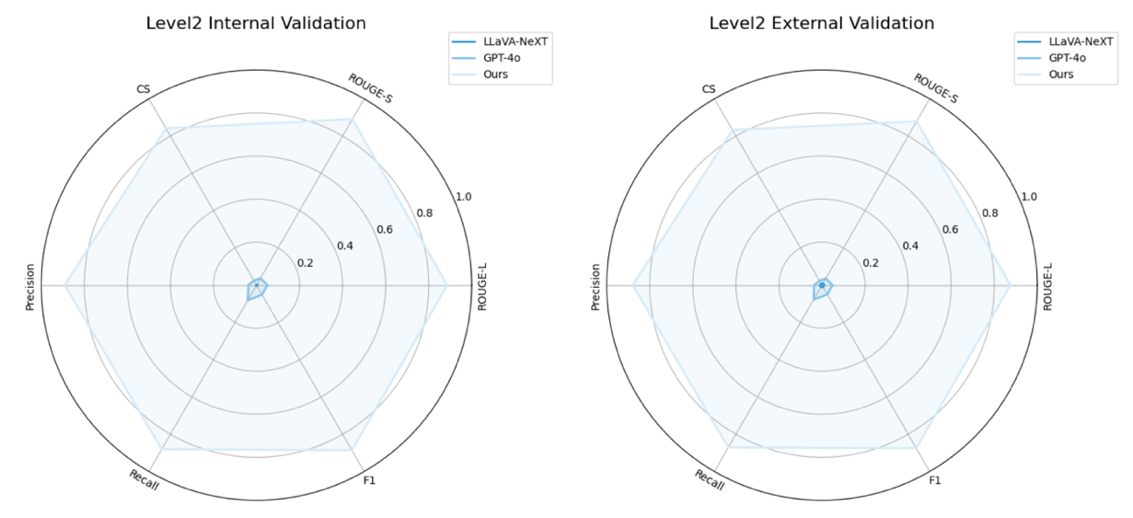}

Figure 3. Overall Performance Comparison on Level 1 and Level 2 Tasks. Our proposed model is evaluated against baseline models (LLaVA-NeXT, ChatGPT-4o) across six metrics (ROUGE-L, ROUGE-S, Cosine Similarity, Precision, Recall, and F1 score). The results indicate that our model consistently and significantly outperforms the baselines on both task levels.

As shown in Figure 3, our proposed LMM significantly outperformed state-of-the-art generalist models, including LLaVA-NeXT and GPT-4o. The performance gap was especially pronounced for the more complex Level 2 tasks compared with the foundational Level 1 tasks. This finding underscores the importance of domain specialization \cite{29,30,31,32} for reliable nuclear medicine image interpretation. The poor performance of LLaVA-NeXT and GPT-4o on Level 2 tasks is attributable to their difficulty in handling clinically critical questions, causing them to repeatedly generate non-diagnostic responses such as canned refusals (e.g., ``I'm unable to interpret medical images... consult a radiologist...'') or produce empty outputs.

\begin{table}[H]
\centering
\scriptsize
\renewcommand{\arraystretch}{0.93}
\setlength{\tabcolsep}{4pt}
\resizebox{\textwidth}{!}{%
\begin{tabular}{@{}lrrrrrr@{}}
\toprule
\textbf{Institution} & \textbf{ROUGE-L} & \textbf{ROUGE-S} & \textbf{Cosine Similarity} & \textbf{Precision} & \textbf{Recall} & \textbf{F1} \\
\midrule
Internal & 0.6239

(± 0.0301) & 0.5644

(± 0.0472) & 0.5369

(± 0.0785) & 0.6168

(± 0.0664) & 0.6647

(± 0.0036) & 0.6323

(± 0.0324) \\
\midrule
External & 0.6389

(± 0.0239) & 0.6257

(± 0.0143) & 0.5287

(± 0.0129) & 0.5782

(± 0.0824) & 0.6371

(± 0.0650) & 0.6648

(± 0.0136) \\
\bottomrule
\end{tabular}%
}
\caption{\footnotesize When analyzing the specific task of identifying metastatic lymph node stations---information encoded in the final five tokens of the model\textquotesingle s output---we observed quantitatively lower scores on both validation sets compared to the holistic performance metrics. Despite this decrease, the model\textquotesingle s ability to detect and localize lymph nodes was high enough to establish its potential clinical utility. These results demonstrate that the model can reliably extract diagnostically relevant information, confirming its feasibility for clinical support.}
\label{tab:5}
\end{table}

Focusing on Level 2 evaluation, our model achieved a primary tumor detection accuracy of 83.14 ± 1.15\% on the internal validation set and 69.03 ± 0.81\% on the external validation set. While the performance on the more challenging task of lymph node localization was comparatively lower (Table 5), it was still substantially higher than that of the generalist baselines, demonstrating the feasibility of building domain-specific models with larger-scale data and model sizes.

\centerimage[width=6.5in,height=3.26042in]{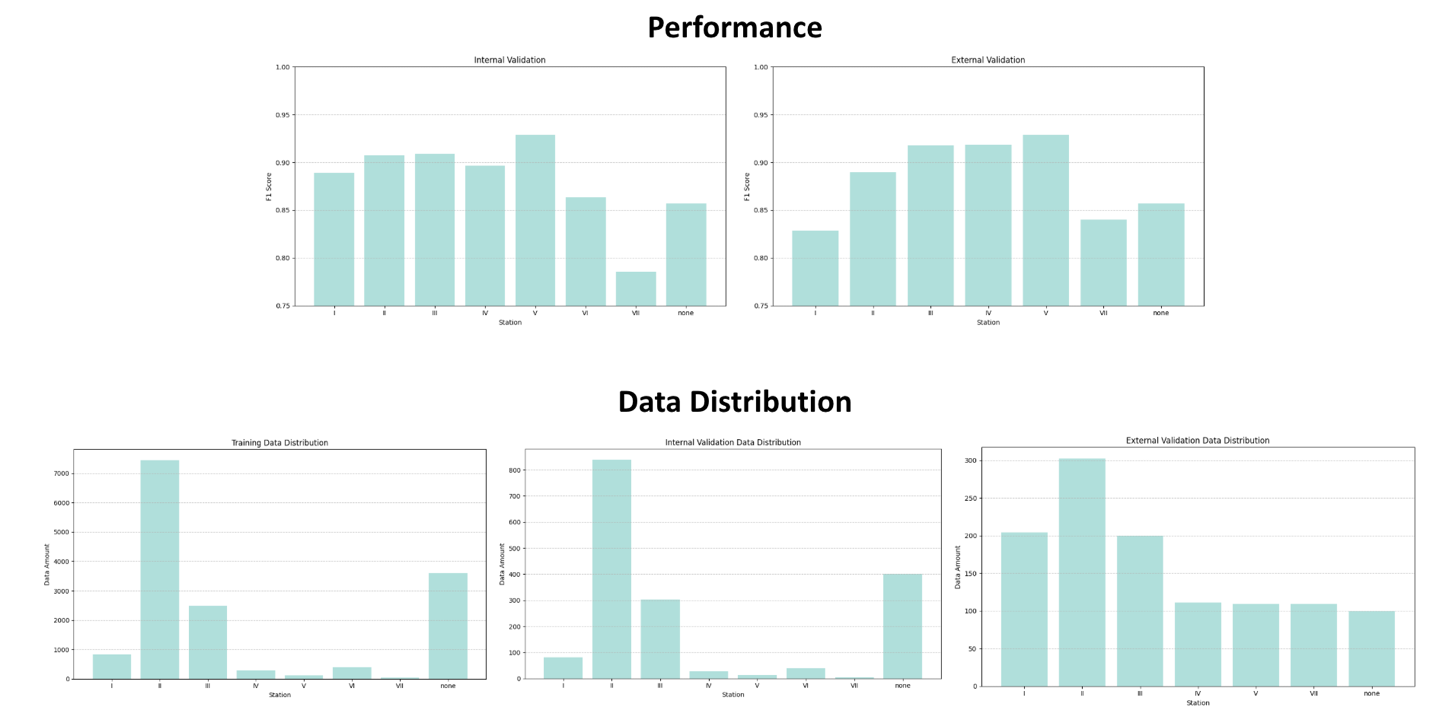}

Figure 4. The top row illustrates the F1 scores for each individual lymph node station, while the bottom row depicts the corresponding data distribution across the training, internal validation, and external validation sets. Notably, despite a significant class imbalance in the training and internal validation data, the model maintains robust and consistent performance across all stations, a result we attribute to the standardized structure of the model\textquotesingle s output.

Figure 4 presents the F1 scores for each lymph node station across both internal and external validation datasets. The model demonstrates remarkably stable and uniform performance across all stations despite the highly imbalanced data distributions. This consistency is likely attributable to the structured output format, which guides the model to focus on classifying clinically significant information rather than on grammatical variations.

Detailed ablation studies covering hyperparameters, pretraining with the PubMed dataset, scalability, and the effect of sentence structure are provided in the Supplementary Notes.

\section{Discussion}

In this study, we developed and evaluated a specialized LMM for interpreting {[}\textsuperscript{18}F{]}FDG PET/CT images in head and neck cancer, addressing the known limitations of generalist models like ChatGPT. Our approach involved fine-tuning the model with a tailored, autoregressive curriculum designed specifically for diagnostic support. We conducted a rigorous quantitative evaluation on a spectrum of tasks, from foundational to clinically critical, and performed a granular analysis by segmenting the model's generated text to assess its performance on diagnostically vital components. As a result, our model significantly outperformed the generalist baselines, highlighting both the necessity and feasibility of developing domain-specific LMMs for clinical applications. While direct qualitative physician assessments were not performed, our quantitative evaluation of clinically salient outputs---such as the detection and localization of primary tumors and lymph nodes---serves as a strong proxy for the model's clinical feasibility.

An alternative strategy is to deploy a suite of smaller specialized experts, such as CNNs or Transformers, each trained for a discrete classification task. While this allows for high performance on narrowly defined objectives, the approach scales poorly, requiring the development and maintenance of a separate model for each clinical question. They are also susceptible to the overfitting. Moreover, in tasks involving free-text generation, such as diagnostic report writing, it is often difficult to define clear-cut rules, making aforementioned expert models impractical to handle the complexity and variability of natural language through rule-based approach.

Despite the advantages, several limitations remain. First, the model demonstrates relatively low performance in extracting clinically relevant information. The model is still susceptible to overfitting and potential security vulnerabilities, emphasizing the need for robust safeguards in clinical applications. Furthermore, the absence of standardized abbreviations in the generated reports reduces clinical readability. The model lacks the ability to fully interpret 3D volumetric data, instead relying on 2D slices, and offers limited interactivity with users, which restricts its utility as a decision-support tool. The training process is computationally intensive and time-consuming. Lastly, our model is currently unable to distinguish the laterality (left vs. right) of the primary tumor and lymph nodes in the sagittal slices.

To address these limitations, several improvements can be considered. First, incorporating domain-specific expert models, such as Transformers or CNNs, as prior can enhance the model\textquotesingle s ability to capture clinically relevant features and improve diagnostic performance. Second, designing an interactive user interface that automatically replaces medical terms with personalized abbreviations could increase the clinical usability and readability of generated reports. Finally, structuring the model's output in a unified and standardized format may not only boost performance but also facilitate more consistent and objective evaluation.

Integrating LMMs specifically tailored to clinical workflows could revolutionize diagnostic support in nuclear medicine. Future research should focus on expanding the training dataset to a more diverse patient population, extending it to additional cancer types and tracers, and validating model performance through prospective, independent studies with nuclear medicine specialists. Furthermore, extending the model to operate directly on 3D volumetric data will help the technology more closely align with real-world clinical needs. These initiatives pave the way for the transition of domain-specific LMMs from proof-of-concept systems to reliable, deployable clinical tools.

\section{Conclusion}

In conclusion, we have developed a specialized LMM to support the diagnosis of head and neck cancer from PET/CT images and demonstrated its feasibility. This was enabled by the construction of a large-scale, multi-institutional dataset curated from seven centers with diverse imaging devices. Our experimental results show that the proposed LMM significantly outperforms state-of-the-art generalist models, such as GPT-4o. These findings indicate that domain-specific LMMs hold strong potential for real-world clinical implementation in diagnostic workflows such as head and neck cancer assessment.

\bibliographystyle{unsrt}
\bibliography{references}

\clearpage
\appendix
\section{Supplementary Materials}

\subsection{Dataset Construction}

To support the development and validation of our model, we meticulously curated three key datasets---HECKTOR2021 \cite{16}, AutoPET2024 \cite{17}, and PubMed---each designed to address different aspects of diagnostic characteristics and educational capabilities in nuclear medicine.

First, the HECKTOR2021 and AutoPET2024 datasets were explicitly curated to support diagnosis assistance. We annotated 3D raw images and then sliced them with preprocessing (standardization, normalization, registration, and resizing) into 256 x 256 x 3 2D slices in sagittal, coronal, and axial directions. The modalities include PET and CT scans, with detailed annotations describing tracer types, image slice orientations, the usage of contrast agents, and the presence or absence of hypermetabolism. Regarding the disease extent, HECKTOR2021 focuses on head and neck cancer cases and includes data from five institutions, each labeled with its respective institution identifier, ensuring a multi-center perspective. AutoPET2024, on the other hand, features data from melanoma, lung cancer, and lymphoma cases, providing a diverse range of disease contexts without institution identifier.

For the level 1 dataset (HECKTOR2021, AutoPET2024), data processing and annotation involved extracting basic and fundamental metadata such as tracer type, imaging direction, modality, anatomical region, the usage of contrast agents, and the presence of hypermetabolism from the raw data from the metadata. The resulting questions are \emph{``Among PET, PET/CT, CT, and MRI, what type of scan is this?'', ``Are there any hypermetabolic lesions or abnormalities in this image?'', ``Among sagittal, coronal, and axial views, what is the view of the image?''}, and ``Was a contrast agent used in this image?''. For the level 2 dataset (HECKHOR2021), two radiologists with different background carefully annotated the existence of the primary tumor, the existence and the location of the lymph node. The resulting question was \emph{``This image is a slice from an FDG PET/CT scan performed on a patient with oropharyngeal cancer. Please assess whether a primary tumor or any cervical lymph nodes suspicious for metastasis are visible in this slice. If any lymph nodes are visible, please specify the station (IL, IR, IIL, IIR, IIIL, IIIR, IVL, IVR, VL, VR, VIL, VIR, VIIL, VIIR, VIIIL, VIIIR, none).''}. These image-conversation (single turn) pairs are converted to JSON format.

Second, the PubMed dataset was curated to enhance the model's general explanation capabilities and to serve as an educational resource. It includes multimodal data such as PET, gamma images, and nuclear medicine-related CT and MRI images, etc. These images are accompanied by broader contextual information derived from their associated open access academic articles. The dataset was constructed by extracting image captions, paragraphs containing descriptions of the images, and their corresponding abstracts and titles. Then, by training ConvNeXT \cite{33}, we filtered the images into nuclear medicine related images or not. Then the nuclear medicine related images' descriptions were converted into an instruction-following format and saved as JSON files.

The final dataset ensures consistency in image size by resizing and padding all images to 256 × 256 pixels. After data preprocessing, the combined datasets were split into training and testing subsets, creating a unified data atlas for our model to handle various diagnostic and educational tasks in nuclear medicine.

\subsection{Ablation Study}

\centerimage[width=6.5in,height=3.72361in]{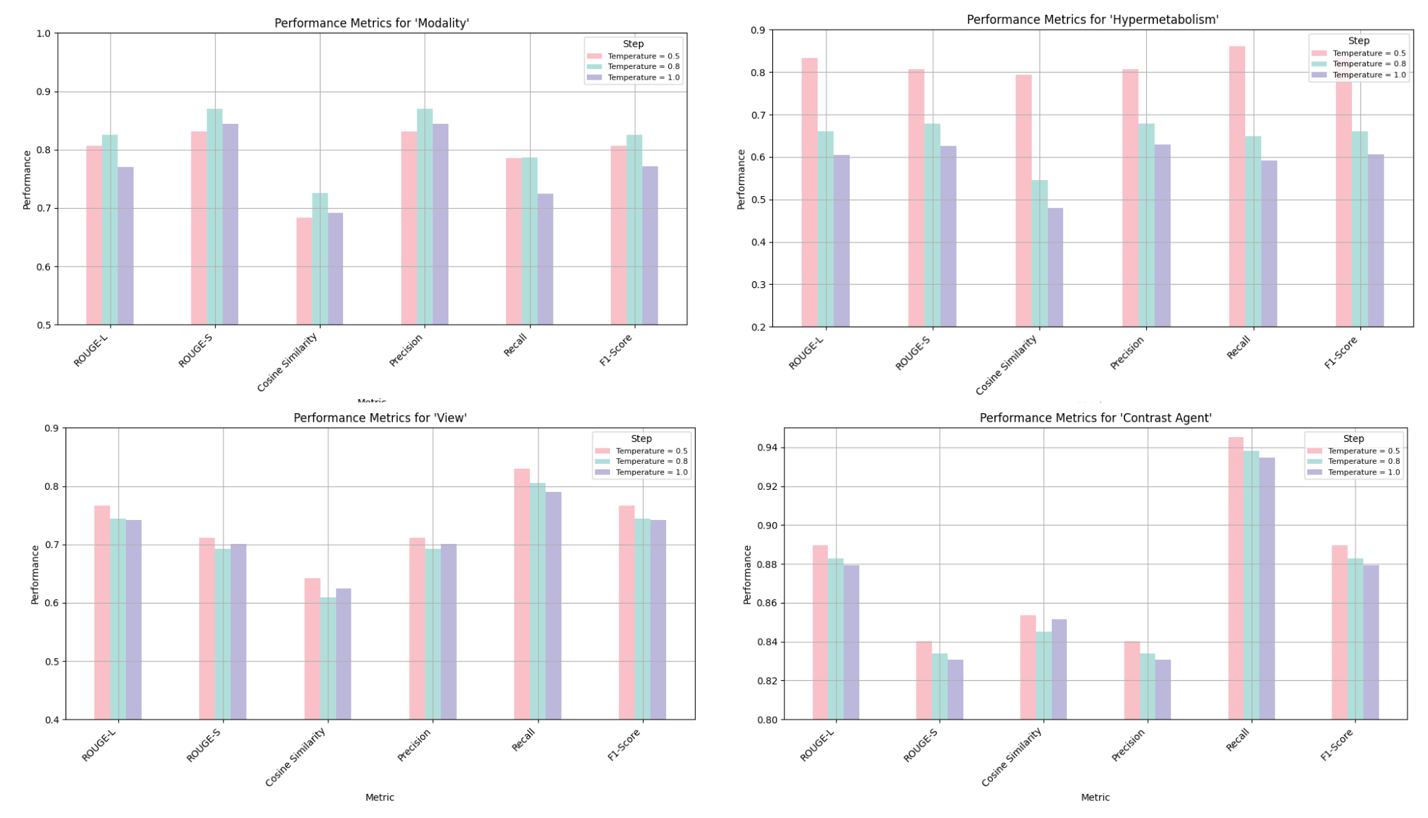}

\centerimage[width=3.23301in,height=1.85449in]{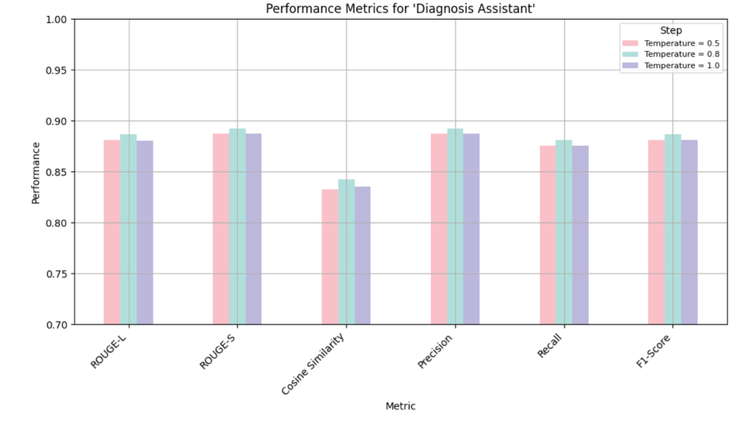}

Figure 5. The performance of diverse metrics on various temperature and top p values such as 0.5 (pink), 0.8 (green), 1.0 (purple). The temperature and top p values are set identically. The performances are evaluated on all question types (modality, existence of hypermetabolism, view, the usage of contrast agent, diagnostic assistance for detecting primary tumor and lymph node) across level 1 and level 2.

We compared the results of various temperature and top p values (0.5, 0.8, 1.0) in the Figure 5. Among those values, we selected 0.8 for the default value because 0.5 generates overly similar answers across all cases although it showed superior performance in the level 1.

\centerimage[width=6.5in,height=3.76111in]{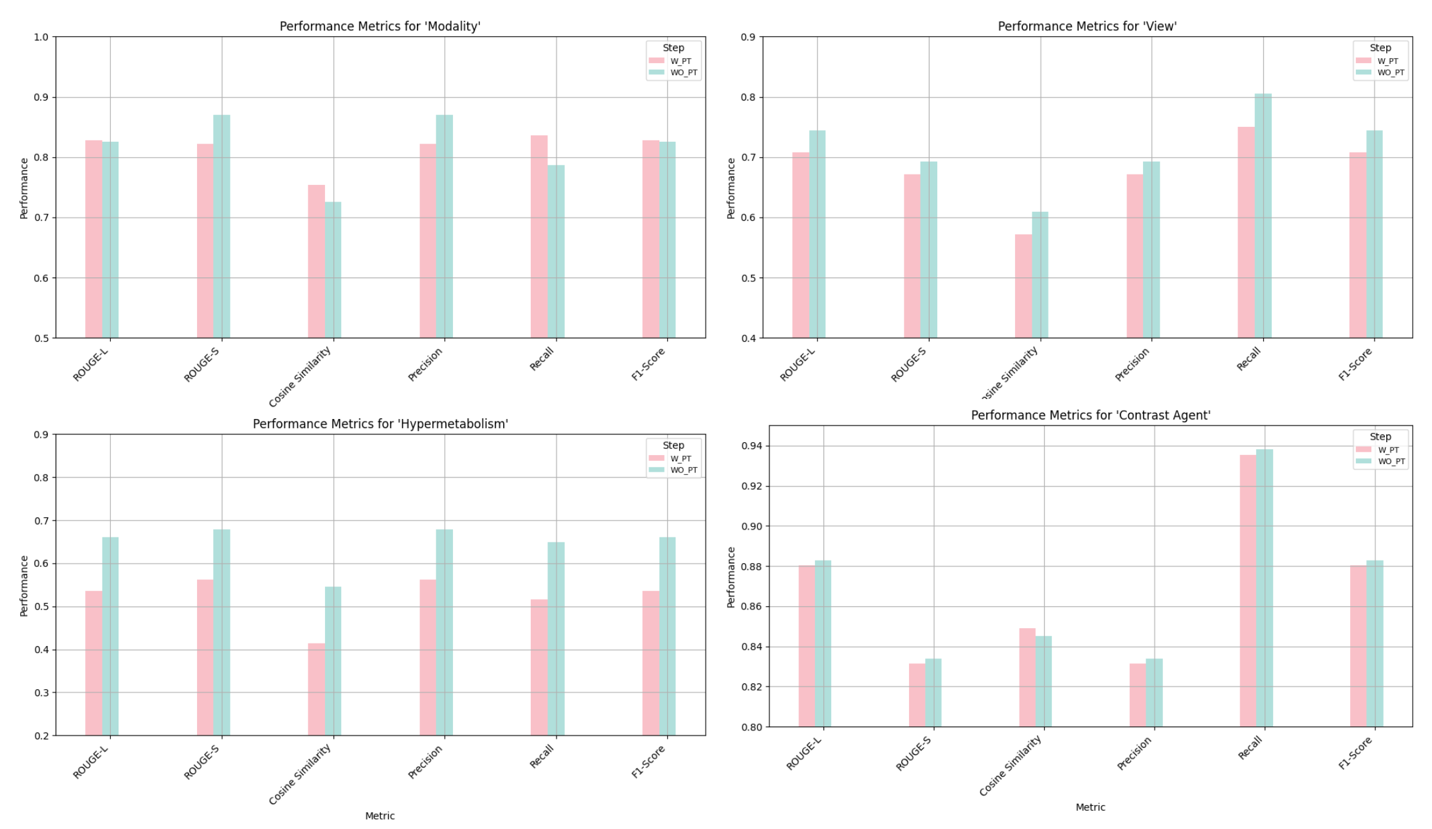}

Figure 6. The results with and without pretraining with PubMed dataset. `W\_PT' (pink) indicates that the performance with the pretraining with PubMed dataset while `WO\_PT' (green) represents the performance without pretraining with PubMed dataset.

To analyze the effect of the pretraining with PubMed dataset, we compared the performance of the internal validation of level 1. As shown in the Figure 6, In most of the cases, the pretraining with PubMed dataset rather degraded the performance. We attribute this to the irrelevance between the information between PubMed's and level 1's questions. In addition, the relatively small size and inherent noise in the PubMed dataset also interfere the model to appropriately learn meaningful knowledges for the downstream tasks.

\centerimage[width=6.5in,height=3.79236in]{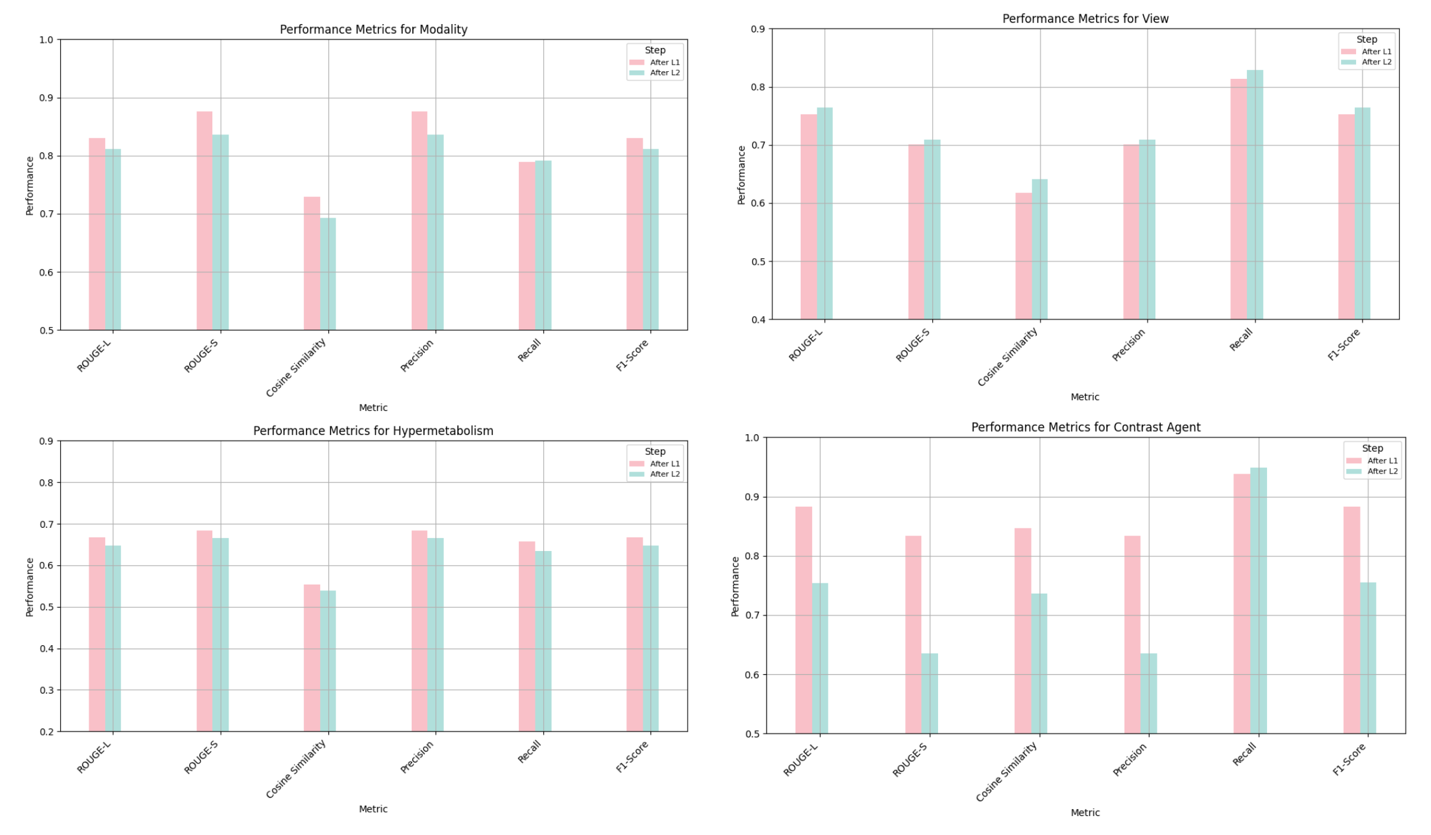}

Figure 7. The change of the level 1 tasks' performances before (pink) and after level 2 training (green).

To test the scalability of the specialized LMM, we compared the level 1 performance before and after training level 2. As demonstrated in Figure 7, after level 2 training, the performance of the level 1 did not significantly perturb in most of the cases showing its versatility.

\centerimage[width=6.5in,height=1.94167in]{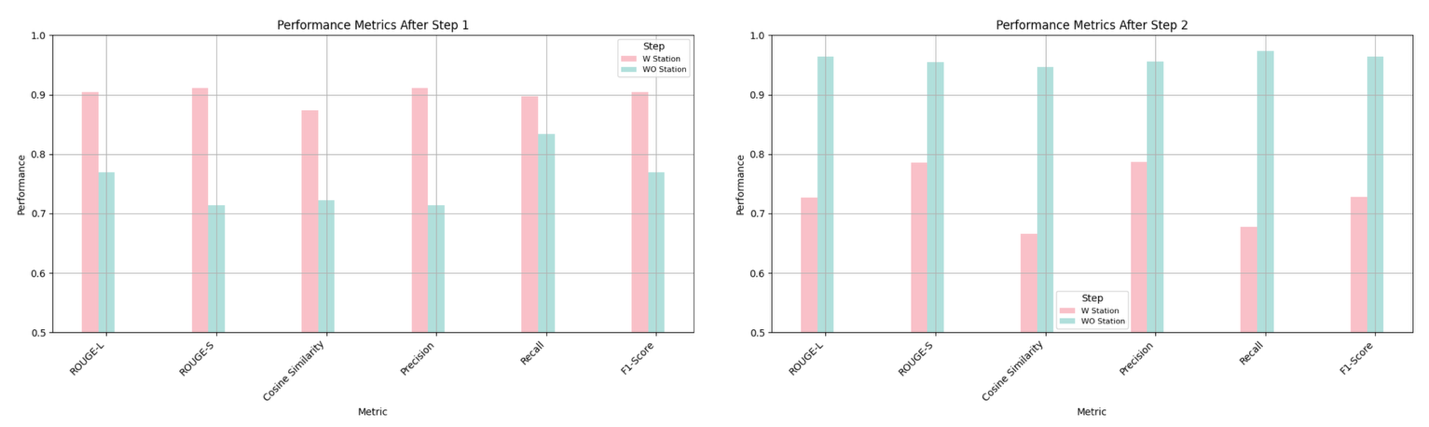}

Figure 8. The change of the performance of answer types, such as with station information (pink) and without station information (green) each of which has different sentence structure.

Figure 8 is about the effect of the answer sentence structure of level 2. We differentiated the sentence structure for the cases with and without lymph node. For example, for the case in which the lymph node does not exist, the answer sentence was ``A primary tumor is visible, but no suspicious cervical lymph nodes are observed.'' while when the lymph node exists, the answer was ``A primary tumor is visible, and suspicious cervical lymph nodes are observed, likely located at station IIL.''. In Figure 8, ``W Station'' indicates the performance of answers with station information while ``WO Station'' represents the performance of answers without station information. In the step 1, we trained model with only ``W Station'' and in the step 2, we trained the model with both 20\% of ``W Station'', and 80 \% of ``WO Station''. As a result, after the step 2, the performance of the ``W Station'' severely degraded. This indicates that when the sentence structure is different in different cases for the same question, it performance can be severely degraded. As a result, we unified the sentence structure in all cases using the abbreviation ``station none'' which indicates the lymph node does not exist.

\subsection{Performance of Level 1 and Level 2}

We evaluated our LMM on three random seeds (42, 84, 200). The scores represent mean and the numbers in the parentheses are standard deviations. Table 6 and 7 are the results of our model on level 1 tasks on internal, and external validations. Table 8 is the performance of our model on level 2 tasks on internal and external validations. As shown in the tables, our model significantly outperforms other baselines proving its feasibility of utilization for diagnostic support.

\begin{table}[H]
\centering
\scriptsize
\renewcommand{\arraystretch}{0.93}
\setlength{\tabcolsep}{4pt}
\resizebox{\textwidth}{!}{%
\begin{tabular}{@{}lrrrrr@{}}
\toprule
\textbf{Model} & \textbf{Modality} & \textbf{Hypermetabolism} & \textbf{Contrast Agent} & \textbf{View} & \textbf{Overall} \\
\midrule
\multicolumn{6}{@{}l@{}}{\textbf{ROUGE-L}} \\
LLaVA-NeXT & 0.046 ± (0.0427) & 0.1433 ± (0.1602) & 0.0688 ± (0.0601) & 0.0385 ± (0.0666) & 0.0741 ± (0.0913) \\
GPT-4o & 0.3714 ± (0.0027) & 0.0999 ± (0.0025) & 0.2228 ± (0.0011) & 0.4597 ± (0.0018) & 0.3044 ± (0.1319) \\
Ours & 0.8088 ± (0.0) & 0.8237 ± (0.0) & 0.8824 ± (0.0) & 0.7726 ± (0.0) & 0.7961 ± (0.0757) \\
\midrule
\multicolumn{6}{@{}l@{}}{\textbf{ROUGE-S}} \\
LLaVA-NeXT & 0.0395 ± (0.0374) & 0.1678 ± (0.0682) & 0.0547 ± (0.0478) & 0.0289 ± (0.05) & 0.0727 ± (0.0731) \\
GPT-4o & 0.2886 ± (0.0024) & 0.0696 ± (0.0018) & 0.1331 ± (0.0008) & 0.3557 ± (0.0019) & 0.2225 ± (0.1089) \\
Ours & 0.8312 ± (0.0) & 0.8 ± (0.0) & 0.8333 ± (0.0) & 0.7174 ± (0.0) & 0.7901 ± (0.0471) \\
\midrule
\multicolumn{6}{@{}l@{}}{\textbf{Cosine Similarity}} \\
LLaVA-NeXT & 0.0486 ± (0.044) & 0.1476 ± (0.1615) & 0.0721 ± (0.0632) & 0.0672 ± (0.1165) & 0.0839 ± (0.0992) \\
GPT-4o & 0.388 ± (0.001) & 0.0891 ± (0.0018) & 0.3022 ± (0.0046) & 0.5374 ± (0.0027) & 0.3372 ± (0.1511) \\
Ours & 0.6848 ± (0.0) & 0.7847 ± (0.0) & 0.8475 ± (0.0) & 0.6497 ± (0.0) & 0.7027 ± (0.1245) \\
\midrule
\multicolumn{6}{@{}l@{}}{\textbf{Precision}} \\
LLaVA-NeXT & 0.0422 ± (0.0395) & 0.1856 ± (0.092) & 0.0601 ± (0.0528) & 0.0324 ± (0.0561) & 0.0801 ± (0.0840) \\
GPT-4o & 0.302 ± (0.0023) & 0.0748 ± (0.0016) & 0.1424 ± (0.0004) & 0.367 ± (0.0011) & 0.2309 ± (0.1107) \\
Ours & 0.8312 ± (0.0) & 0.8 ± (0.0) & 0.8333 ± (0.0) & 0.7174 ± (0.0) & 0.7901 ± (0.0471) \\
\midrule
\multicolumn{6}{@{}l@{}}{\textbf{Recall}} \\
LLaVA-NeXT & 0.0855 ± (0.0779) & 0.2507 ± (0.3314) & 0.132 ± (0.115) & 0.0762 ± (0.132) & 0.1361 ± (0.1786) \\
GPT-4o & 0.6953 ± (0.0037) & 0.1961 ± (0.0029) & 0.7646 ± (0.0032) & 0.8743 ± (0.0051) & 0.6553 ± (0.2453) \\
Ours & 0.7892 ± (0.0) & 0.85 ± (0.0) & 0.9375 ± (0.0) & 0.837 ± (0.0) & 0.8128 ± (0.1154) \\
\midrule
\multicolumn{6}{@{}l@{}}{\textbf{F1}} \\
LLaVA-NeXT & 0.0497 ± (0.0459) & 0.1675 ± (0.1959) & 0.076 ± (0.0667) & 0.0432 ± (0.0748) & 0.0841 ± (0.1090) \\
GPT-4o & 0.3891 ± (0.0028) & 0.1074 ± (0.0022) & 0.2383 ± (0.0004) & 0.4772 ± (0.0007) & 0.3168 ± (0.1342) \\
Ours & 0.8088 ± (0.0) & 0.8237 ± (0.0) & 0.8824 ± (0.0) & 0.7726 ± (0.0) & 0.7961 ± (0.0757) \\
\bottomrule
\end{tabular}%
}
\caption{\footnotesize The performance of internal validation of level 1 tasks.}
\label{tab:6}
\end{table}

\begin{table}[H]
\centering
\scriptsize
\renewcommand{\arraystretch}{0.93}
\setlength{\tabcolsep}{4pt}
\resizebox{\textwidth}{!}{%
\begin{tabular}{@{}lrrrr@{}}
\toprule
\textbf{Model} & \textbf{Modality} & \textbf{Hypermetabolism} & \textbf{View} & \textbf{Overall} \\
\midrule
\multicolumn{5}{@{}l@{}}{\textbf{ROUGE-L}} \\
LLaVA-NeXT & 0.0468 ± (0.0046) & 0.084 ± (0.0153) & 0.0846 ± (0.0097) & 0.0718 ± (0.0209) \\
GPT-4o & 0.3769 ± (0.0034) & 0.0901 ± (0.003) & 0.4797 ± (0.005) & 0.3285 ± (0.1510) \\
Ours & 0.8117 ± (0.0) & 0.7211 ± (0.0) & 0.7692 ± (0.0) & 0.7673 ± (0.0393) \\
\midrule
\multicolumn{5}{@{}l@{}}{\textbf{ROUGE-S}} \\
LLaVA-NeXT & 0.0403 ± (0.0046) & 0.1289 ± (0.0456) & 0.0635 ± (0.0071) & 0.0776 ± (0.0460) \\
GPT-4o & 0.3021 ± (0.0034) & 0.0627 ± (0.002) & 0.3761 ± (0.0057) & 0.2532 ± (0.1215) \\
Ours & 0.8346 ± (0.0) & 0.7 ± (0.0) & 0.7143 ± (0.0) & 0.7496 ± (0.0640) \\
\midrule
\multicolumn{5}{@{}l@{}}{\textbf{Cosine Similarity}} \\
LLaVA-NeXT & 0.0492 ± (0.0079) & 0.0911 ± (0.0222) & 0.1452 ± (0.0217) & 0.0952 ± (0.0447) \\
GPT-4o & 0.3822 ± (0.0018) & 0.0791 ± (0.0023) & 0.5548 ± (0.0036) & 0.3449 ± (0.1782) \\
Ours & 0.6888 ± (0.0) & 0.642 ± (0.0) & 0.6434 ± (0.0) & 0.6581 ± (0.0231) \\
\midrule
\multicolumn{5}{@{}l@{}}{\textbf{Precision}} \\
LLaVA-NeXT & 0.0426 ± (0.0055) & 0.1368 ± (0.0428) & 0.0722 ± (0.0103) & 0.0839 ± (0.0472) \\
GPT-4o & 0.3164 ± (0.004) & 0.0669 ± (0.0016) & 0.3859 ± (0.0063) & 0.2609 ± (0.1242) \\
Ours & 0.8346 ± (0.0) & 0.7 ± (0.0) & 0.7143 ± (0.0) & 0.7496 ± (0.0640) \\
\midrule
\multicolumn{5}{@{}l@{}}{\textbf{Recall}} \\
LLaVA-NeXT & 0.0859 ± (0.0131) & 0.1201 ± (0.0375) & 0.1753 ± (0.0297) & 0.1271 ± (0.0463) \\
GPT-4o & 0.6689 ± (0.0021) & 0.1759 ± (0.006) & 0.883 ± (0.0049) & 0.6144 ± (0.2767) \\
Ours & 0.7917 ± (0.0) & 0.7444 ± (0.0) & 0.8333 ± (0.0) & 0.7898 ± (0.0385) \\
\midrule
\multicolumn{5}{@{}l@{}}{\textbf{F1}} \\
LLaVA-NeXT & 0.0502 ± (0.0056) & 0.0939 ± (0.0187) & 0.0962 ± (0.0129) & 0.0801 ± (0.0253) \\
GPT-4o & 0.3959 ± (0.0038) & 0.0961 ± (0.0026) & 0.4949 ± (0.0058) & 0.3396 ± (0.1546) \\
Ours & 0.8117 ± (0.0) & 0.7211 ± (0.0) & 0.7692 ± (0.0) & 0.7673 ± (0.0393) \\
\bottomrule
\end{tabular}%
}
\caption{\footnotesize The performance of external validation of level 1.}
\label{tab:7}
\end{table}

\begin{table}[H]
\centering
\scriptsize
\renewcommand{\arraystretch}{0.93}
\setlength{\tabcolsep}{4pt}
\resizebox{\textwidth}{!}{%
\begin{tabular}{@{}lrrrrrr@{}}
\toprule
\textbf{Model} & \textbf{ROUGE-L} & \textbf{ROUGE-S} & \textbf{Cosine Similarity} & \textbf{Precision} & \textbf{Recall} & \textbf{F1} \\
\midrule
\multicolumn{7}{@{}l@{}}{\textbf{Internal Validation}} \\
LLaVA-NeXT & 0.0026 ± (0.0000) & 0.0025 ± (0.0000) & 0.0014 ± (0.0000) & 0.0025 ± (0.0000) & 0.0028 ± (0.0000) & 0.0026 ± (0.0000) \\
GPT-4o & 0.0503 ± (0.0009) & 0.0373 ± (0.0007) & 0.0243 ± (0.0006) & 0.0379 ± (0.0008) & 0.0810 ± (0.0022) & 0.0512 ± (0.0012) \\
Ours & \textbf{0.8856 ± (0.0000)} & \textbf{0.8912 ± (0.0000)} & \textbf{0.8413 ± (0.0000)} & \textbf{0.8912 ± (0.0000)} & \textbf{0.8804 ± (0.0000)} & \textbf{0.8856 ± (0.0000)} \\
\midrule
\multicolumn{7}{@{}l@{}}{\textbf{External Validation}} \\
LLaVA-NeXT & 0.0093 ± (0.0001) & 0.0096 ± (0.0001) & 0.0069 ± (0.0001) & 0.0108 ± (0.0001) & 0.0103 ± (0.0001) & 0.0103 ± (0.0001) \\
GPT-4o & 0.0486 ± (0.0012) & 0.0361 ± (0.0008) & 0.0235 ± (0.0006) & 0.0366 ± (0.0007) & 0.0779 ± (0.0018) & 0.0494 ± (0.0011) \\
Ours & \textbf{0.8751 ± (0.0000)} & \textbf{0.8794 ± (0.0000)} & \textbf{0.8324 ± (0.0000)} & \textbf{0.8794 ± (0.0000)} & \textbf{0.8711 ± (0.0000)} & \textbf{0.8751 ± (0.0000)} \\
\bottomrule
\end{tabular}%
}
\caption{\footnotesize Internal and external validation of level 2.}
\label{tab:8}
\end{table}

\end{document}